\documentclass[10pt,twocolumn]{article}

\usepackage[margin=0.75in,top=1.0in,bottom=1.0in]{geometry}
\usepackage{times}
\usepackage{microtype}
\usepackage{amsmath,amssymb,amsfonts}

\usepackage{graphicx}
\usepackage{booktabs}
\usepackage{multirow}
\usepackage{caption}
\usepackage{subcaption}
\usepackage{array}

\usepackage{xcolor}
\usepackage{enumitem}
\usepackage[colorlinks=true,linkcolor=blue!60!black,
            citecolor=blue!60!black,urlcolor=blue!60!black]{hyperref}
\usepackage{url}

\usepackage{fancyhdr}
\usepackage{titlesec}
\titleformat{\section}{\large\bfseries}{\thesection.}{0.5em}{}
\titleformat{\subsection}{\normalsize\bfseries}{\thesubsection}{0.5em}{}
\titleformat{\subsubsection}{\normalsize\itshape\bfseries}{\thesubsubsection}{0.5em}{}
\titlespacing{\section}{0pt}{10pt}{4pt}
\titlespacing{\subsection}{0pt}{7pt}{2pt}
\titlespacing{\subsubsection}{0pt}{5pt}{2pt}

\usepackage{tikz}
\usetikzlibrary{shapes,arrows,positioning,fit,backgrounds,calc}

\usepackage{pifont}
\newcommand{\cmark}{\ding{51}}
\newcommand{\xmark}{\ding{55}}

\usepackage[numbers,sort&compress]{natbib}
\usepackage{placeins}
\usepackage{dblfloatfix}
\usepackage{afterpage}

\begin{document}

\twocolumn[{%
\begin{@twocolumnfalse}
\vspace*{4pt}
\begin{center}
  {\LARGE\bfseries Act--Observe--Rewrite:\\[5pt]
   Multimodal Coding Agents as In-Context\\[5pt]
   Policy Learners for Robot Manipulation}\\[14pt]
  {\large Vaishak Kumar}\\[4pt]
  {\normalsize\textit{General Bionix, Inc.}}\\[12pt]
\end{center}

\noindent\rule{\textwidth}{0.4pt}
\vspace{2pt}

\noindent\textbf{Abstract.}\quad
Can a multimodal language model learn to manipulate physical objects by
reasoning about its own failures---without gradient updates, demonstrations, or
reward engineering? We argue the answer is yes, under conditions we characterise
precisely. We present \textbf{Act--Observe--Rewrite (AOR)}, a framework in which
an LLM agent improves a robot manipulation policy by synthesising entirely new
executable Python controller code between trials, guided by visual observations
and structured episode outcomes. Unlike prior work that grounds LLMs in
pre-defined skill libraries or uses code generation for one-shot plan synthesis,
AOR makes the full low-level motor control implementation the unit of LLM
reasoning, enabling the agent to change not just what the robot does, but how it
does it. The central claim is that interpretable code as the policy representation creates
a qualitatively different kind of in-context learning from opaque neural policies:
the agent can \emph{diagnose} systematic failures and \emph{rewrite} their causes.
We validate this across three robosuite manipulation tasks and report promising
results, with the agent achieving high success rates without demonstrations,
reward engineering, or gradient updates.

\vspace{6pt}
\noindent\rule{\textwidth}{0.4pt}
\vspace{8pt}
\end{@twocolumnfalse}
}]

\section{Introduction}

Robot manipulation is entering an era of \emph{foundation model integration}.
Vision-language-action models trained on internet-scale data can generalise
across hundreds of tasks zero-shot \citep{brohan2023rt2,black2024pi0}. Yet two
significant challenges persist: (1)~when these models fail on a specific
deployment configuration, diagnosing \emph{why} and adapting \emph{without
retraining} remains largely unsolved; and (2)~the sheer cost of pretraining
data and compute places such models out of reach for researchers who want to
iterate quickly on novel manipulation tasks.

A complementary tradition has emerged in language agent research. Reflexion
\citep{shinn2023reflexion} showed that an LLM can improve performance on
decision-making and programming tasks purely through verbal self-reflection---
storing episode summaries in memory and using them as additional context in
subsequent trials, with no weight updates. ReAct \citep{yao2023react}
demonstrated that interleaving reasoning traces with environment actions
significantly outperforms imitation or RL baselines in grounded settings.
Voyager \citep{wang2023voyager} built an open-ended lifelong agent in Minecraft
whose skill library grows entirely from LLM-generated executable code.

These results suggest a different paradigm for robot learning: instead of
training a policy, \emph{write} one---and rewrite it when it fails. The appeal
is clear: no data collection, no training loops, full interpretability, and
direct application of the LLM's world knowledge about geometry, physics, and
programming. The question is whether this paradigm is viable for continuous
physical manipulation, where success depends on precise kinematics, noisy
vision, workspace geometry, and object physics---none of which admit the
kind of clean natural-language characterisation that makes verbal reflection
effective in text-based environments.

This paper addresses that question directly. We introduce
\textbf{Act--Observe--Rewrite (AOR)}, a two-timescale framework:
the robot executes an episode under a Python controller;
between episodes, a multimodal LLM reasons over key-frame images and structured
outcome data, diagnoses failure modes, and outputs an entirely new controller
class. The new controller is dynamically compiled, validated, and loaded for
the next trial.

The central claim of this paper is not a performance number.
It is an \emph{architectural claim}: that making the full controller
implementation the unit of LLM reasoning---rather than a skill selector, a
parameter vector, or a reward function---qualitatively changes what can be
learned in-context. Specifically:
\begin{itemize}[leftmargin=*,itemsep=1pt,topsep=2pt]
\item The LLM can \emph{diagnose the cause} of a failure in terms of code logic
  (``the y-coordinate is negated in the back-projection, which is wrong for
  OpenGL convention''), not just observe that failure occurred.
\item The LLM can make \emph{architectural changes}---adding phases, changing
  the target computation, introducing geometric corrections---that go far beyond
  parameter tuning.
\item The LLM can reason about \emph{physical constraints}---workspace limits,
  sensor biases, grip geometry---and encode that reasoning as executable
  corrections.
\end{itemize}

\paragraph{Contribution:}
\textbf{The AOR framework}: a general architecture for code-synthesis
reflexive learning in physical manipulation. Implemented with Claude Code
as the coding agent, AOR achieves 100\% on two tasks and 91\% on a third.
The experiments are reported honestly: the residual failures on Stack are
an observed shortcoming---the agent identified gripper contact with the
target cube as the cause but failed to discover a placement strategy that
avoids it, even though one likely exists.

\section{Background and Problem Formulation}

\subsection{The In-Context Policy Learning Problem}

We consider a robot manipulation agent operating in episodic trials. At each
timestep $t$, the agent observes images $o_t$ and proprioceptive state $s_t$,
and produces an action $a_t \in \mathcal{A}$. A trial terminates with outcome
$\tau = \{r, T, \phi_1,\ldots,\phi_k\}$ where $r$ is the scalar reward, $T$ is
the duration, and $\phi_i$ are structured diagnostic signals (phase log,
minimum distance to target, oscillation flag).

The \emph{in-context policy learning problem} is: given the history of $n$
past outcomes $H_n = \{\tau_1, \ldots, \tau_n\}$ and a set of key-frame images
$\mathcal{I}_n$, produce an improved policy $\pi_{n+1}$ without updating any
model weights. Prior approaches differ in how they represent $\pi$:

\begin{itemize}[leftmargin=*,itemsep=1pt,topsep=2pt]
\item \textbf{Parameter update}: $\pi$ is a fixed architecture; in-context
  improvement adjusts a parameter vector $\theta$ \citep{shinn2023reflexion}.
\item \textbf{Skill selection}: $\pi$ selects from a fixed skill library
  $\mathcal{S}$; in-context improvement changes the selection policy
  \citep{ahn2022saycan,huang2022innermonologue}.
\item \textbf{Reward synthesis}: $\pi$ is trained by RL; in-context
  improvement rewrites the reward function
  \citep{ma2024eureka}.
\item \textbf{Code synthesis} (AOR): $\pi$ is an executable Python class;
  in-context improvement rewrites the entire implementation.
\end{itemize}

The code synthesis approach is the most expressive: it can represent all of the
above plus arbitrary algorithmic changes, new phases, and new geometric
computations. Its risk is that the hypothesis space is so large that an LLM may
produce incorrect or unsafe code. AOR addresses this with a compilation sandbox,
action clamping, and a fallback to the previous working controller.

\section{Related Work}
\label{sec:related}

\subsection{Verbal Self-Reflection for Agents}

Reflexion \citep{shinn2023reflexion} is the closest antecedent to AOR. It
introduced the principle of verbal reinforcement: after each trial, the agent
produces a natural-language reflection stored in a persistent memory buffer,
which is prepended to the context for the next trial. Reflexion demonstrated
22\% improvement on AlfWorld and 11\% on HumanEval. However, Reflexion was
designed for \emph{text-based} environments (symbolic planning, code
generation) where observations are structured and actions are discrete. The
reflections modify \emph{strategy descriptions} in natural language, not
executable motor-control code.

AOR extends Reflexion to physical manipulation in three ways: (i)~observations
include raw RGB-D images requiring visual reasoning; (ii)~the policy is
executable motor-control code, not a language description; (iii)~failure modes
involve continuous kinematics and sensor geometry, not just logical errors.
The key insight from Reflexion that AOR preserves and generalises is that
\emph{verbalising failure modes in natural language before rewriting policy is
more effective than blind parameter search}.

Hong et al.\ \citep{hong2024reflective} propose Reflective Test-Time Planning
for embodied LLMs, combining reflection-in-action (pre-execution candidate
generation), reflection-on-action (post-execution updates), and retrospective
reflection (long-horizon credit assignment). Their work is the most direct
embodied parallel to Reflexion and demonstrates strong improvements on
long-horizon household tasks and MuJoCo environments. Feng et al.\
\citep{feng2025reflective} extend this line with a VLM-based framework that
imagines future world states to guide action selection across multi-stage
manipulation tasks, outperforming MCTS and commercial VLMs on long-horizon
benchmarks. \textbf{The key distinction from AOR} is the level of abstraction:
both reflective planning approaches operate at the level of \emph{sub-goal
planning}---deciding which pre-built skill to invoke next---while AOR operates
at the level of \emph{skill implementation}, rewriting the motor-control code
that executes a skill. The approaches are thus complementary rather than
competing: one could use reflective planning for sub-goal selection while AOR
handles low-level controller adaptation.

Self-Refine \citep{madaan2023selfrefine} generalises the reflection idea to
arbitrary generation tasks: the same LLM produces output, critiques it, then
refines it iteratively, improving code optimisation, maths, and dialogue quality.
The generate--feedback--refine loop in AOR is structurally identical, but grounded
in a physical domain: the ``feedback'' is visual and kinematic evidence from an
episode rollout, not LLM self-critique, and the ``output'' is executable
motor-control code, not free-form text.

ReAct \citep{yao2023react} interleaved reasoning traces with environment
interactions, achieving 34\% absolute gains over RL/imitation baselines on
AlfWorld with 1--2 in-context examples. AdaPlanner \citep{sun2023adaplanner}
added in-plan and out-of-plan revision strategies, reducing samples required by
2$\times$. These works demonstrate strong in-context planning but, like
Reflexion, operate in symbolic environments with discrete action spaces.

\subsection{Code Generation for Robot Control}

Code as Policies \citep{liang2023cap} demonstrated that an LLM can generate
hierarchically structured robot control programs from natural language
commands. Given a task description and a library of perception primitives, the
LLM produces Python code that calls these primitives to implement the task.
\textbf{The key distinction from AOR}: Code as Policies is \emph{one-shot}---it
generates code once and executes it. There is no closed-loop improvement from
episode outcomes, no visual evidence of failure, and no iterative refinement.
AOR adds the critical feedback loop: observe what happened, reason about why,
rewrite.

ProgPrompt \citep{singh2023progprompt} used Pythonic program specifications to
ground LLM task plans in available robot actions, achieving state-of-the-art on
VirtualHome. Like Code as Policies, it is one-shot at the plan level, though
the program representation is more structured.

SayPlan \citep{rana2023sayplan} and related works use LLMs to generate
scene-graph-grounded plans from natural language. These approaches focus on
\emph{what to do} (task planning), not \emph{how to do it} (motor-control
implementation). AOR targets the latter.

ReKep \citep{huang2024rekep} uses VLMs to express manipulation constraints as
Python functions over tracked 3D keypoints, enabling zero-shot multi-stage and
bimanual tasks from a single natural-language instruction. This is the closest
prior work to AOR's representation: both write Python code that reasons over
visual observations at runtime. \textbf{The key distinction}: ReKep generates
constraint code once from a scene description; it does not improve that code
across episodes. AOR explicitly accumulates failure evidence across trials
and rewrites the code in response, enabling systematic correction of errors
that are invisible in a single observation.

\subsection{LLM-Generated Reward Functions and Fitness Criteria}

Eureka \citep{ma2024eureka} used GPT-4 to generate reward function code,
optimised over many RL training runs via an evolutionary outer loop, achieving
human-expert-level reward design on 83\% of 29 RL environments including
dexterous tasks such as pen spinning. DrEureka \citep{ma2024dreureka} extended
this to automate sim-to-real domain randomisation. Language to Rewards
\citep{yu2023languagerewards} translates natural-language task descriptions into
reward function parameters for MuJoCo locomotion and manipulation controllers,
bridging semantic instruction and low-level optimisation without manual reward
engineering. RLCD \citep{yang2023rlcd} and related works use LLM feedback as a
substitute for human reward labelling.

\textbf{The key distinction from AOR}: Eureka-style methods require many full
RL training runs as the inner loop. Each outer-loop LLM call is followed by
thousands of gradient updates. AOR requires only episode execution---each
LLM call is followed by a single trial. This makes AOR substantially cheaper
and faster for rapid task adaptation, at the cost of lower peak performance on
tasks that benefit from large-scale RL.

The two approaches are complementary: Eureka is optimal when RL training is
feasible and the goal is maximum performance; AOR is optimal when fast
adaptation and interpretability matter more than peak performance.

\subsection{LLM-Based Optimisation}

A growing body of work uses LLMs as black-box optimisers: rather than
computing a gradient, the LLM is shown a history of (solution, score) pairs
and asked to propose a better solution.

OPRO \citep{yang2024opro} formalised this idea directly: a meta-prompt
contains past solutions sorted by score, and the LLM is asked to generate
new candidates to maximise the objective.  On prompt engineering and linear
regression tasks OPRO matched or exceeded human-crafted solutions using only
natural-language feedback.  FunSearch \citep{romera2024funsearch} applied the
same principle to combinatorial mathematics: an evolutionary outer loop
selects high-scoring programs and feeds them back to the LLM as context,
discovering new results in extremal graph theory and bin-packing.
EvoPrompting \citep{chen2024evoprompting} used LLMs to propose neural
architecture mutations, combining evolutionary search with in-context program
synthesis.

\textbf{AOR is an instance of LLM-based optimisation} where the objective is
episode success rate, the solution space is Python controller code, and the
``score'' is a rich multimodal signal --- key-frame images, reward traces, and
phase-transition logs --- rather than a scalar.  The critical difference from
OPRO and FunSearch is that AOR's feedback is \emph{visual and causal}: the
LLM can see \emph{where} the arm was at the moment of failure and reason about
the physical cause, not merely observe that the score was low.  This richer
feedback enables targeted single-step improvements (e.g.\ fixing a sign error
in the back-projection formula) that pure score-based optimisation would
require many trials to discover by coincidence.

\subsection{Grounded Planning with LLMs}

SayCan \citep{ahn2022saycan} grounded high-level LLM planning with a learned
value function scoring which robot skills are executable in the current scene.
PaLM-SayCan achieved 74\% execution success on long-horizon manipulation tasks.
The key limitation is that SayCan requires pre-trained low-level skills;
it cannot improve those skills through reflection.

Inner Monologue \citep{huang2022innermonologue} continuously injected
environment feedback as a running ``inner monologue''---passive scene
descriptions and success detection---into the LLM prompt, enabling closed-loop
replanning. Inner Monologue improved on SayCan by allowing the LLM to detect
and recover from execution failures, but its feedback is high-level (``the
object was not grasped''), not diagnostic (``the arm descended 6\,cm above the
object because the depth centroid is on the visible surface, not the centre'').
AOR's key contribution over Inner Monologue is the \emph{depth} of diagnosis:
by reasoning over images and controller code simultaneously, the LLM can
pinpoint root causes at the level of individual lines of code.

GPT-4V for Robotics \citep{wake2024gpt4v} uses multimodal LLMs to analyse
human demonstration videos and synthesise robot programs. This is complementary
to AOR: it provides a rich warm-start policy from demonstrations; AOR then
refines it through self-improvement.

\subsection{Foundation Models and VLAs for Manipulation}

The dominant paradigm for generalised robot manipulation trains
\emph{vision-language-action (VLA) models} jointly on internet-scale data and
robot trajectories. RT-2 \citep{brohan2023rt2} co-fine-tuned a large VLM on
robot data, achieving emergent generalisation. PaLM-E \citep{driess2023palme}
injected continuous sensor modalities into a 562B-parameter language model.
OpenVLA \citep{kim2024openvla} provided an open-source 7B-parameter alternative
that outperforms RT-2-X (55B) by 16.5\% with LoRA fine-tuning. $\pi_0$
\citep{black2024pi0} combined a VLM backbone with a flow-matching action expert
for dexterous manipulation. Octo \citep{ghosh2024octo} introduced the first
open-source generalist policy trained on 800k episodes from Open X-Embodiment,
supporting both language and goal-image conditioning with strong cross-embodiment
transfer. The Open X-Embodiment \citep{oxe2023} dataset and Gato
\citep{reed2022gato} established the multi-task, multi-embodiment training
paradigm. Most recently, dual-system architectures such as GR00T N1
\citep{nvidia2025groot} mirror cognitive fast/slow thinking: a slow VLM
(System~2) reasons about instructions and scene context while a fast visuomotor
policy (System~1) produces real-time actions. This two-system structure echoes
AOR's two-timescale design, but the timescale separation in GR00T N1 operates
\emph{within} a single inference pass, whereas AOR's slow loop operates
\emph{between episodes}, accumulating experience to rewrite the fast controller.

\textbf{AOR occupies a fundamentally different position from VLA models} along
several axes (Table~\ref{tab:comparison}). VLA models optimise for
\emph{breadth}: they generalise across hundreds of tasks through pretraining.
AOR optimises for \emph{depth and interpretability}: it achieves high
performance on individual tasks through targeted reasoning, with a policy
that can be read, audited, and modified by a human. The two are complementary:
AOR-style reflection could provide targeted debugging when a VLA model fails
on a specific deployment configuration without requiring full retraining.

Diffusion Policy \citep{chi2023diffusion} and ACT \citep{zhao2023act} represent
the state of the art in data-driven visuomotor policy learning. Diffusion Policy
outperforms prior methods by 46.9\% on 12 tasks using multi-modal action
distributions; ACT achieves 80--90\% on bimanual tasks from 10 minutes of
demonstrations. These establish an important comparison: \emph{given
demonstrations, data-driven methods outperform AOR}. The contribution of AOR
is performance \emph{without any demonstrations}.

\subsection{In-Context Learning and Adaptation in Robotics}

Instant Policy \citep{vosylius2025instant} demonstrated in-context imitation
learning from 1--2 demonstrations via graph diffusion over scene graphs,
enabling zero-shot generalisation to new tasks without weight updates. This is
the closest gradient-free test-time adaptation baseline to AOR; the key
distinction is that Instant Policy adapts to a \emph{new task} from a
demonstration, whereas AOR improves on the \emph{same task} through its own
failure history without any demonstrations.

RoboAgent \citep{bharadhwaj2023roboagent} achieved generalisation across 38
tasks from 7,500 demonstrations by combining semantic augmentation with
multi-task action-chunking. These approaches remain fundamentally
demonstration-driven; AOR requires no demonstrations.

The approach most structurally similar to AOR at the code level is the
concurrent work of Kagaya et al.\ on memory-augmented code adaptation
\citep{kagaya2025mtp}, which maintains a library of successful control programs
from prior environments and retrieves and adapts them as in-context examples for
new settings. The key distinction is directionality: that work retrieves
\emph{successes} to prime new tasks, whereas AOR diagnoses \emph{failures}---
using visual keyframes as evidence---to iteratively correct the same task's
controller across episodes.

Closest in spirit to AOR otherwise is Eureka's use of LLM-generated code with
a feedback loop---but as argued above, the feedback loop in Eureka involves RL
training, not episode-level trial and error. To our knowledge, AOR is the
first system to demonstrate iterative \emph{controller implementation}
improvement for robot manipulation via LLM reflection over visual evidence,
without any gradient updates.

\begin{table*}[t]
\centering
\caption{Comparison of AOR to related approaches along key dimensions. ``Code unit'' refers
to whether the code generated is a full controller implementation, a reward
function, a plan, or a skill-selection policy. ``Failure-image FB'' indicates
whether the system uses visual key-frames from failed episodes as feedback.}
\label{tab:comparison}
\small
\setlength{\tabcolsep}{3.5pt}
\begin{tabular}{lcccccccc}
\toprule
Approach & Visual input & Code unit & Iterative & Failure-image FB & Zero-shot & Interpretable & Physical \\
\midrule
Reflexion \citep{shinn2023reflexion} & \xmark & Strategy & \cmark & \xmark & \cmark & \cmark & \xmark \\
Self-Refine \citep{madaan2023selfrefine} & \xmark & General & \cmark & \xmark & \cmark & \cmark & \xmark \\
Reflective TTP \citep{hong2024reflective} & \cmark & Sub-goal & \cmark & \xmark & \cmark & Partial & \cmark \\
Code as Policies \citep{liang2023cap} & \xmark & Full plan & \xmark & \xmark & \cmark & \cmark & \cmark \\
ReKep \citep{huang2024rekep} & \cmark & Constraint & \xmark & \xmark & \cmark & \cmark & \cmark \\
Eureka \citep{ma2024eureka} & \xmark & Reward fn & \cmark & \xmark & \cmark & \cmark & \cmark \\
Inner Monologue \citep{huang2022innermonologue} & \xmark & Skill select & \cmark & \xmark & \cmark & Partial & \cmark \\
Instant Policy \citep{vosylius2025instant} & \cmark & Action & \xmark & \xmark & \cmark & \xmark & \cmark \\
RT-2 / OpenVLA / Octo & \cmark & Action & \xmark & \xmark & \cmark & \xmark & \cmark \\
Diffusion Policy / ACT & \cmark & Action & \xmark & \xmark & \xmark & \xmark & \cmark \\
\midrule
\textbf{AOR (this work)} & \cmark & \textbf{Full ctrl} & \cmark & \cmark & \cmark & \cmark & \cmark \\
\bottomrule
\end{tabular}
\end{table*}

\section{The AOR Framework}
\label{sec:framework}

\begin{figure*}[!t]
\centering
\begin{tikzpicture}[
  graybox/.style={rectangle, rounded corners=3pt,
                  draw=gray!55, fill=gray!7,
                  text width=2.2cm, align=center,
                  minimum height=0.95cm, font=\small},
  greenbox/.style={rectangle, rounded corners=3pt,
                   draw=green!50!black, fill=green!5,
                   text width=2.2cm, align=center,
                   minimum height=0.95cm, font=\small},
  bluebox/.style={rectangle, rounded corners=4pt,
                  draw=blue!55, fill=blue!5,
                  text width=3.4cm, align=center,
                  minimum height=1.15cm, font=\small},
  orangebox/.style={rectangle, rounded corners=4pt,
                    draw=orange!65, fill=orange!5,
                    text width=3.4cm, align=center,
                    minimum height=1.15cm, font=\small},
  garr/.style={->, >=stealth, thick, gray!60},
  barr/.style={->, >=stealth, thick},
  rewrite/.style={->, >=stealth, dashed, line width=1.2pt, blue!65},
  lbl/.style={font=\footnotesize\itshape, text=gray!55, inner sep=2pt}
]

\node[graybox]  (cam)  {RGB+Depth\\Camera};
\node[greenbox, right=1.2cm of cam]  (vis)  {Vision\\Pipeline};
\node[greenbox, right=0.7cm of vis]  (feat) {Feature\\Dict};
\node[greenbox, right=0.7cm of feat] (ctrl) {Controller\\(Python class)};
\node[graybox,  right=1.2cm of ctrl] (env)  {Simulation\\/ Robot};

\begin{scope}[on background layer]
  \node[draw=green!50!black, line width=1.2pt, fill=green!3,
        rounded corners=6pt, inner xsep=8pt, inner ysep=6pt,
        fit=(vis)(feat)(ctrl),
        label={[font=\small\bfseries, text=green!40!black]above:%
               Policy\,---\,rewritten by LLM each episode}]
       (policy) {};
\end{scope}

\draw[garr] (cam.east)  -- (vis.west);   
\draw[barr] (vis.east)  -- (feat.west);  
\draw[barr] (feat.east) -- (ctrl.west);  
\draw[garr] (ctrl.east) -- (env.west);   

\node[bluebox,   below=2.5cm of feat] (hla)
  {Multimodal LLM\\[3pt]\textit{Diagnose} failure\\\textit{Rewrite} Policy};
\node[orangebox, below=2.5cm of env]  (mem)
  {Episodic Memory\\[2pt]reward, phases,\\key-frame images};

\draw[barr] (env.south) --
  node[lbl, right]{episode end}
  (mem.north);

\draw[barr] (mem.west) --
  node[lbl, above]{images + outcomes}
  (hla.east);

\draw[rewrite] (hla.north) --
  node[lbl, left]{rewrite}
  (policy.south);

\end{tikzpicture}
\caption{The AOR two-timescale loop.
The \textbf{Policy} (green box)---comprising the vision pipeline, feature
extraction, and Python controller---is the unit synthesised by the LLM after
each episode.
\textit{Fast loop} (top): sensor observations flow through the Policy to produce
actions each timestep, repeating at control frequency throughout an episode.
\textit{Slow loop} (bottom): at episode end, outcomes and key-frame images enter
episodic memory; the multimodal LLM diagnoses failure modes and synthesises an
entirely new Policy, dynamically compiled for the next trial.
No model weights are updated at any point.}
\label{fig:architecture}
\end{figure*}
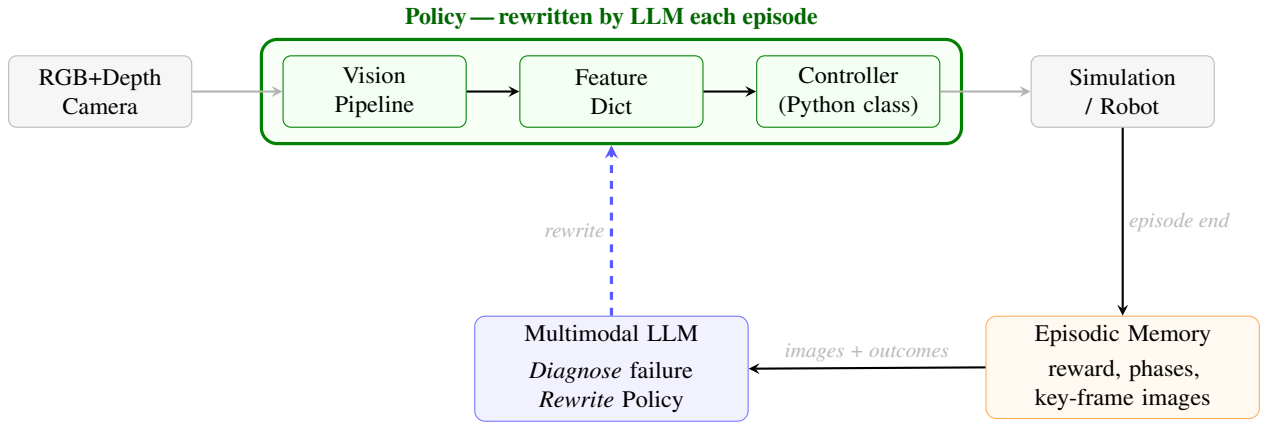

AOR instantiates a two-timescale loop illustrated in Figure~\ref{fig:architecture}.
The \emph{fast loop} runs within an episode at control frequency; the
\emph{slow loop} runs between episodes at reasoning frequency. This
decomposition is a deliberate design choice motivated by three considerations.

\textbf{Separation of concerns.} Physical control requires low-latency
computation with deterministic timing guarantees. LLM inference is slow,
expensive, and non-deterministic. Keeping the LLM strictly between episodes
preserves real-time control properties while maximising the quality of the
between-episode reasoning.

\textbf{Episode-level abstraction.} Single-timestep observations are too noisy
for reliable LLM reasoning about manipulation failures. Episode-level summaries
(phase logs, key frames, reward trajectories) abstract away within-episode noise
and expose the systematic patterns that the LLM needs to reason about.

\textbf{Code as the interface.} The Python \texttt{Controller} class is a
natural interface between LLM reasoning and robot execution: it is human-readable
(supporting diagnosis), machine-executable (supporting deployment), and
compositionally modifiable (supporting targeted rewrites).

\subsection{The Four Components}

\paragraph{Vision pipeline:} Converts RGB-D images into a feature dict
$\mathbf{f}_t = \{$\texttt{object\_pos}, \texttt{eef\_pos},
\texttt{detected}, \ldots$\}$ at each timestep. The vision pipeline is
deliberately separate from the controller so that the LLM can reason about and
rewrite them independently. The instantiation uses HSV colour-based segmentation, with the
camera convention corrections described in Section~\ref{sec:instantiation}.
\paragraph{Controller:} A Python class with \texttt{reset()} and
\texttt{get\_action(features)} methods, safety-clamped via a compilation
sandbox. The controller can implement any strategy expressible in Python with
NumPy: phase-based state machines, PID loops, waypoint sequences, geometric
computations. This expressiveness is the key advantage of code synthesis over
parameter-tuning or skill-selection approaches.
\paragraph{Episodic memory:} Stores structured episode outcomes (reward, step
count, phase-transition log, minimum EEF-to-object distance, oscillation flag)
and a small set of key-frame images captured at phase transitions and at regular
intervals. The memory is persistent across sessions.
\paragraph{Multimodal LLM agent:} Receives the full memory context plus the
current controller source code and produces a new controller class. The agent is
prompted to answer three questions before producing code: (1)~What was the
dominant failure mode? (2)~Was the root cause in vision, controller logic, or
parameters? (3)~What is the single most impactful change? This structured
reflection is directly inspired by Reflexion's \citep{shinn2023reflexion} verbal
reinforcement principle but instantiated for code synthesis.

\subsection{Safety and Stability Mechanisms}

Several mechanisms prevent the code synthesis loop from producing unsafe or
degenerate behaviour:

\textbf{Compilation sandbox.} All generated code is compiled and validated in an
isolated namespace before execution. Any code that fails to compile, lacks the
required interface, or raises exceptions at instantiation falls back to the
previous working controller.

\textbf{Action clamping.} Every action returned by the controller is
safety-clamped: $[\text{-}2, 2]$ for Cartesian deltas, $[\text{-}1, 1]$ for the
gripper. Runtime exceptions per step produce a zero (safe-stop) action.

\textbf{Bounded rewrites.} The LLM is instructed to make targeted changes
addressing the diagnosed failure mode, preserving the overall structure of the
previous controller. Holistic rewrites that abandon all prior knowledge are
discouraged. This is analogous to the bounded update principle in Reflexion.

\section{Instantiation for Robot Manipulation}
\label{sec:instantiation}

AOR is instantiated for the robosuite simulation platform. The key technical
choices that implement the abstract framework described in
Section~\ref{sec:framework} are presented here; experimental details follow in Section~\ref{sec:exp}.

Importantly, the design choices described in this section were \emph{not}
prescribed by the authors. The agent was given only a task description,
a robosuite environment, and a camera feed. Every architectural decision---the
choice of colour segmentation over learned detection, the phase-based
controller structure, the EMA smoothing scheme, the grasp-retry logic---emerged
from the agent's own iterative diagnosis and rewriting across episodes.
The authors observed and recorded these decisions; they did not make them.

\subsection{Vision Pipeline Design Choices}

\subsubsection{Colour segmentation and the camera convention problem}

For tasks with known-colour objects, the agent's vision pipeline converged on
HSV colour segmentation. A key insight discovered through AOR's reflexive loop concerns
\emph{camera convention consistency}. Robosuite uses
\texttt{IMAGE\_CONVENTION = ``opengl''}, meaning row 0 is at the image bottom
and the camera uses $y$-up, $z$-back axes. The correct pinhole back-projection is:
\begin{equation}
  \mathbf{p}_{\mathrm{cam}} = [x_p,\; y_p,\; {-}d,\; 1]^\top,\quad
  x_p = \tfrac{(u-c_x)d}{f_x},\quad y_p = \tfrac{(v-c_y)d}{f_y}.
  \label{eq:backproject}
\end{equation}
The standard OpenCV formula uses $-y_p$, producing 5--8\,cm errors. A further
subtlety is that robosuite's \texttt{get\_camera\_extrinsic\_matrix} applies an
OpenCV axis correction incompatible with OpenGL images; the world-to-camera
transform must be computed directly from \texttt{cam\_xmat}. These bugs are
subtle, undocumented, and were autonomously discovered by the AOR reflexive
loop---not by manual inspection (Section~\ref{sec:perception_debug}).

\subsection{Controller Representation}

The default controller template implements a sequential phase-based state
machine: $\texttt{reach} \to \texttt{descend} \to \texttt{grasp} \to
\texttt{lift} \to \{\text{task-specific phases}\}$. Phase transitions are
threshold-based; each phase has a timeout fallback.

Two design principles emerged as invariant across all instantiations through the
agent's iterative reasoning:

\textbf{Stationary grasp.} During gripper closure, the EEF holds its position.
Any downward pressing force destabilises the object. This is geometrically
obvious but was not in the default controller; the LLM identified it as the
root cause of grasp failure in the first Lift iteration.

\textbf{EMA smoothing.} Actions are smoothed via $\mathbf{a}_t = \alpha
\hat{\mathbf{a}}_t + (1{-}\alpha)\mathbf{a}_{t-1}$ with $\alpha \in [0.3, 0.5]$.
The appropriate $\alpha$ depends on object geometry: larger, more inertially
stable objects tolerate higher $\alpha$; lighter objects require
higher smoothing to maintain grip integrity.

\section{Experimental Validation}
\label{sec:exp}

\subsection{Setup}

All experiments run on robosuite v1.5.1 under MuJoCo 3.2.3 with a UR5e arm
and OSC\_POSE Cartesian delta controller. The coding agent is
\textbf{Claude Code} (claude-sonnet-4-x family), Anthropic's agentic coding
assistant, which handles multimodal episode analysis and controller rewriting.
All visual observations come from a single 256$\times$256 RGB-D
\texttt{agentview} camera; ground-truth object positions are never used in the
controller.

\paragraph{Agent comparison:}
We also ran the same three tasks using OpenAI's Codex agent (GPT~5.3, High
reasoning setting) as a direct substitute for Claude Code. The Codex agent was
unable to solve any of the three tasks within the same iteration budget,
failing to produce a working controller even for the simplest Lift task. This
result highlights that agent capability is a meaningful variable in the AOR
loop, and that the framework's effectiveness is not independent of the
underlying coding agent.

\paragraph{Tasks:}
Figure~\ref{fig:tasks} shows the three tasks. We use three robosuite tasks of
increasing complexity to test the claims in Section~\ref{sec:framework}:

\textbf{Lift}: Pick up a red cube from a table surface. A diagnostic
task designed to surface vision calibration requirements and validate
systematic-failure rapid convergence. Success is unambiguous; failures
produce informative phase logs.

\textbf{PickPlaceCan}: Pick up a cola can and place it into a designated
bin. Tests whether AOR can resolve a systematic perception failure
requiring correct colour identification and robust blob selection---specifically,
the can renders as \emph{red} (not silver) in the agentview camera, and the
bin's own red indicator marker must be suppressed.

\textbf{Stack}: Pick up a red cubeA and place it stacked on green cubeB.
Requires two sequential visually-guided operations with $\sim$2\,cm placement
tolerance. The hardest task: exposes both systematic vision pipeline bugs and
stochastic grasp dynamics, exercising multiple categories of the tractability
taxonomy across 20 controller versions.

\afterpage{%
\begin{figure*}[p]
\centering
\setlength{\tabcolsep}{3pt}
\renewcommand{\arraystretch}{0.5}
\begin{tabular}{ccc}
  \includegraphics[width=0.315\textwidth]{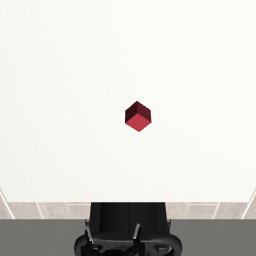} &
  \includegraphics[width=0.315\textwidth]{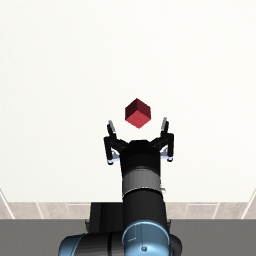} &
  \includegraphics[width=0.315\textwidth]{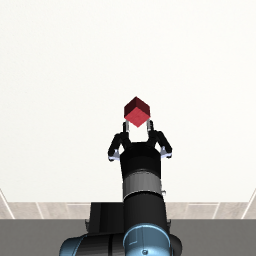} \\[2pt]
  \includegraphics[width=0.315\textwidth]{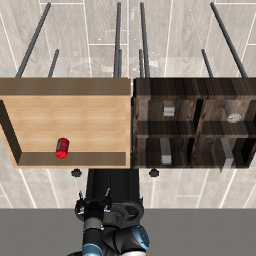} &
  \includegraphics[width=0.315\textwidth]{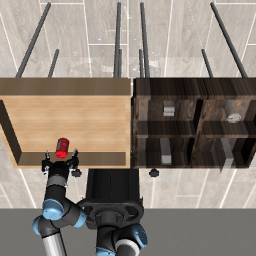} &
  \includegraphics[width=0.315\textwidth]{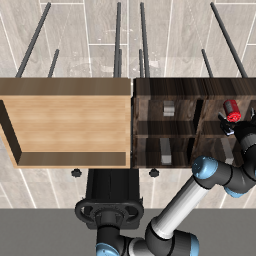} \\[2pt]
  \includegraphics[width=0.315\textwidth]{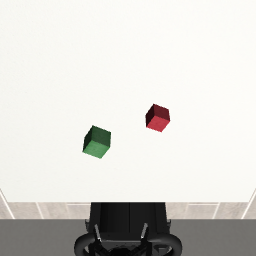} &
  \includegraphics[width=0.315\textwidth]{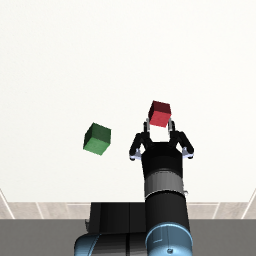} &
  \includegraphics[width=0.315\textwidth]{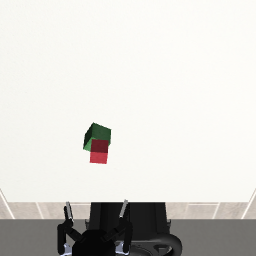} \\[4pt]
  \small Initial scene & \small Grasp & \small Completion \\
\end{tabular}
\caption{The three evaluation tasks, each shown at the initial scene, grasp, and
task completion (all from the final controller version). \emph{Top row:}
\textbf{Lift} --- pick up the red cube; converges to 100\% in 3 LLM calls.
\emph{Middle row:} \textbf{PickPlaceCan} --- grasp the cola can and drop it
into the bin; achieves 100\% after diagnosing the camera colour pipeline.
\emph{Bottom row:} \textbf{Stack} --- pick up the red cube and place it on the
green cube; reaches 91\% after 20 controller versions addressing both vision
pipeline bugs and stochastic grasp dynamics.
All observations are from the 256$\times$256 \texttt{agentview} camera;
no ground-truth object positions are used.}
\label{fig:tasks}
\end{figure*}
\clearpage
}

\subsection{Experimental Observations by Failure Type}
\label{sec:perception_debug}

The experiments reveal a consistent pattern: AOR's behaviour tracks the nature
of the underlying failure mode. Evidence for each observed category follows.

\paragraph{Systematic failures: rapid convergence}

\textbf{Lift} converged to 100\% in 3 LLM calls. Episode\,0 failed entirely:
the phase log showed \texttt{final\_phase: reach} at 500/500 steps, indicating
a systematic bias preventing descend. The LLM diagnosed a 2.5\,cm vision depth
bias and introduced a $z$-offset correction. Episode\,1 reached grasp but
failed: images showed the gripper flanking the cube and then the cube on the
table---the LLM diagnosed downward pressing during closure as the root cause
and changed to a stationary grasp. Episode\,2 succeeded; 4/4 randomised
repeats also succeeded.

\textbf{PickPlaceCan} converged to 100\% in 2 LLM calls, each resolving
one of two compounding perception bugs.

\emph{Call~1 --- wrong object colour.}
Episode\,0 showed the arm stalled in the reach phase for all 1000 steps.
The LLM diagnosed a total detection failure: the initial
\texttt{detect\_silver\_can} routine (low-saturation HSV mask) found no
detections because the can renders as \emph{red}, not metallic silver, in
the agentview camera. The vision branch was rewritten to use a red-hue
segmentor.

\emph{Call~2 --- bin indicator contamination.}
With red detection active, Episode\,1 showed the arm descending toward the
\emph{bin} side of the scene rather than the can on the table.
The LLM diagnosed: \texttt{detect\_red\_object} averages all red pixels in
the frame; the robosuite bin area contains a small red colour marker
indicating the can's target slot, whose pixels pull the centroid rightward.
The fix was to switch to largest-connected-component filtering
(\texttt{detect\_red\_can}), isolating the can---the largest red region---
from the smaller bin indicator.

Episode\,2 succeeded; 20/20 randomised repeats also succeeded, all without
any simulator object-position readout.

\textbf{Stack (v0--v10)} exposed systematic vision pipeline bugs invisible at
Lift's precision tolerance but producing 5--8\,cm errors at Stack's 2\,cm
placement tolerance. After six controller iterations failed to improve placement
precision, the LLM identified \emph{vision inaccuracy} as the root cause.
Two bugs were found: (i)~back-projection with $-y_p$ instead of $+y_p$
(wrong for OpenGL convention); (ii)~extrinsic matrix with an OpenCV axis
correction incompatible with OpenGL images. Correcting both brought
localisation errors to 1.4\,cm (cubeA) / 1.9\,cm (cubeB). Four further
controller iterations reached 70\% (Table~\ref{tab:stack_progress}).

\begin{table}[h]
\centering
\caption{Stack controller progression. v0--v9 fix vision and control bugs;
v17--v20 add grasp retries; residual failures are gripper--cubeB contact.}
\label{tab:stack_progress}
\small
\setlength{\tabcolsep}{4pt}
\begin{tabular}{llc}
\toprule
Version & Key change & Success \\
\midrule
v0--v6  & Controller tuning (vision error 5--8\,cm) & 0\% \\
v7--v8  & Back-projection y-flip fix                & 33\% \\
v9      & Extrinsic convention fix                  & 45\% \\
v10     & Calibrated vision                         & 70\% \\
\midrule
v11--v16 & Retract ramp, descend tuning             & $\sim$75\% \\
v17     & Grasp retry on failure                    & 6/10 \\
v18     & Variable approach offset (buggy)          & 76\% \\
v19     & Separate offset per cube                  & 83\% \\
\textbf{v20} & \textbf{Freeze cubeB at place entry} & \textbf{91\%} \\
\bottomrule
\end{tabular}
\end{table}

\paragraph{Observed shortcoming: failure to find a working placement strategy}

The remaining 9\% failure rate in Stack after v20 is not stochastic noise.
Every failure episode completes the full phase sequence---grasp, lift, align,
place, release, retract---but the stack does not hold. The cause is consistent
and visible in episode images: the gripper fingers contact cubeB during the
descent to place cubeA, physically displacing it before release.

The agent observed this correctly across failure episodes. After v20, the LLM's
reflection identified gripper contact with cubeB as the root cause. However,
it did not discover a placement strategy that avoids the contact. Approaches
such as a steeper descent angle, a compliant release from slightly above, or a
lateral nudge-and-release likely exist and could resolve the issue---the agent
did not explore them. The iteration terminated without a fix.

This is an honest shortcoming of the reflexion search as run in these experiments.
The agent correctly diagnosed the failure but got stuck: it could not construct,
within the hypothesis space it explored, a controller variant that places cubeA
without disturbing cubeB. Whether a sufficiently capable agent or a longer
search would find the solution is an open question.

\subsection{Overall Performance Summary}

Table~\ref{tab:main} summarises peak performance and iteration counts across
all three tasks, annotated with the dominant failure type observed.

\begin{table}[h]
\centering
\caption{Peak success rate and LLM-call count. Stack residual failures involve
gripper contact with cubeB during placement; the agent diagnosed this but did
not find a fix.}
\label{tab:main}
\resizebox{\columnwidth}{!}{%
\begin{tabular}{lccl}
\toprule
Task & Success & LLM calls & Residual failure \\
\midrule
Lift         & \textbf{100\%} &  3 & None \\
PickPlaceCan & \textbf{100\%} &  2 & None \\
Stack        & \textbf{91\%}  & 20 & Gripper contact (unresolved) \\
\bottomrule
\end{tabular}%
}
\end{table}

\section{Discussion}

\subsection{Code as the Right Representation for Reflective Learning}

The central question of this paper was whether making the full controller
implementation the unit of LLM reasoning qualitatively changes in-context
learning in physical manipulation. The evidence says yes.

Consider the Stack task vision bugs. In a parameter-tuning framework
(Reflexion-style), the LLM would observe that placement precision is poor and
reduce thresholds or adjust gains. These adjustments cannot fix a sign error in
the back-projection formula. In a skill-selection framework (SayCan,
Inner Monologue), the skill implementations are fixed; the LLM cannot reach
into them. In Code as Policies, there is no feedback loop to trigger
re-examination. Only a code-synthesis framework with a closed loop and
diagnostic prompts can identify ``the y-coordinate in the back-projection has
the wrong sign'' and rewrite it. This is a fundamental capability advantage,
not just a quantitative improvement.

The colour-pipeline diagnosis in PickPlaceCan makes the same point: the
episode log showed the arm stuck in the reach phase for all 1000 steps,
indicating a total detection failure. Recognising that the can rendered as
\emph{red} rather than silver, and that averaging all red pixels produced a
centroid biased toward the bin's colour marker, required understanding what the
segmentor was computing and why. No parameter tuning or skill selection can
express this; it requires rewriting the object-localisation logic.

\subsection{Autonomous Perception Debugging}

The autonomous discovery of the camera convention bugs deserves explicit
discussion, as it illustrates a form of debugging absent from conventional
robot learning. In standard imitation or RL pipelines, vision bugs are
invisible: the policy simply fails to train to high performance, and the
attribution to vision vs.\ controller is unclear.

In AOR, the LLM has access to both the observation data (images showing where
the arm went) and the implementation code (the back-projection formula).
After Stack iteration 6, the LLM produced a diagnosis of the form: ``the
controller changes have improved convergence speed but the placement is
consistently wrong by 5--8\,cm---this is larger than any achievable
parameter-tuning effect and the magnitude is consistent with a systematic
back-projection error.'' This level of attribution---from symptom to code
module to specific line---is only possible because the LLM can read and reason
about the code implementation.

This suggests that code-synthesis reflexive agents may be useful not just as
policy learners but as \emph{debugging tools} for any vision-based robot system,
even ones that ultimately use learned policies.

\subsection{Comparison to the Reflexion Family}

The most precise comparison is with Reflexion \citep{shinn2023reflexion}
and Reflective Test-Time Planning \citep{hong2024reflective}. All three
operate in the in-context improvement paradigm; the differences are
architectural.

Reflexion modifies \emph{verbal strategies}, which are then interpreted by
a high-level LLM producing discrete actions. This is effective in symbolic
environments but cannot produce the geometric computations (back-projection
correction, bin geometry target correction) needed for precise manipulation.

Reflective Test-Time Planning \citep{hong2024reflective} operates at the
sub-goal level: it decides which pre-built primitive to invoke next and how
to adapt the plan when a sub-goal fails. The primitives themselves are fixed.
AOR is complementary: one could use Reflective Test-Time Planning to select
\emph{which task to execute} while AOR learns \emph{how to execute it}.

The generalisable principle emerging from all three works is:
\emph{in-context improvement is effective when the level of abstraction at
which the LLM operates matches the level at which failures occur}. Reflexion
matches text-level planning failures; Reflective TTP matches sub-goal-level
failures; AOR matches motor-control implementation failures.

\section{Limitations and Future Work}

\textbf{Simulation only.} All results are from simulation. Real robots introduce
sensing noise, actuation imprecision, and lighting variation that would require
tighter uncertainty modelling. The techniques the agent introduced
(colour-based segmentation, camera convention correction, grasp-retry
mechanisms) are motivated by physical principles and should transfer to real
hardware.

\textbf{Sequential iteration.} Each hypothesis is tested one trial at a time.
Parallel hypothesis testing---generating multiple candidate controllers and
evaluating them simultaneously---would reduce wall-clock time substantially.

\textbf{LLM-dependence.} The quality of diagnosis depends on the LLM's
capability to reason about geometry, code, and physics. As LLMs improve, AOR
performance is expected to improve without any changes to the framework.

\textbf{Incomplete search.} The Stack experiments reveal a shortcoming of the
reflexion search: the agent correctly identified gripper contact with the target
cube as the cause of residual failures but exhausted its iteration budget without
finding a placement strategy to avoid it. The agent converged to a local optimum
in controller space. Mechanisms such as explicit diversity pressure, restarts, or
structured exploration of placement strategies are not present in the current
AOR loop and would be needed to overcome this kind of search failure.

\textbf{Classical controller bias.} Left unprompted, the agent converged on
classical control techniques---HSV segmentation, phase-based state machines,
PID-style gains---because these are well-represented in its training data and
produce interpretable, debuggable code. The agent is not inherently restricted
to classical methods: prompting it to use a specific technique (learned visual
features, neural controllers, Kalman filtering) would likely cause it to adopt
that approach instead. The choice of technique is a prompt-level decision, and
more modern perception or control methods could be introduced simply by
instructing the agent to use them. Exploring this prompt-steerable technique
selection is a direct avenue for future work.

\paragraph{Future directions:}
(i)~Prompting the agent to use modern perception methods (e.g.\ learned keypoint
detectors or foundation-model visual features) in place of hand-coded HSV
segmentation, to improve robustness to lighting variation and novel objects.
(ii)~Combining AOR with a VLA prior for warm-started adaptation.
(iii)~Using Reflective Test-Time Planning \citep{hong2024reflective} for
sub-goal selection while AOR handles low-level implementation adaptation.
(iv)~Real-robot deployment with appropriate uncertainty modelling.
(v)~Theoretical characterisation of when in-context policy improvement
converges, in terms of observability and reproducibility of failure modes.

\section{Conclusion}

This paper presented Act--Observe--Rewrite, a framework for in-context policy
learning in robot manipulation that makes executable Python controller code the
unit of LLM reasoning. The framework advances beyond prior work in verbal reflection
(Reflexion, Reflective Test-Time Planning) by operating at the level of
\emph{motor-control implementation} rather than strategy description or sub-goal
selection, and beyond prior work in robot code generation (Code as Policies,
ProgPrompt) by closing the feedback loop with visual episode evidence.

On Lift and PickPlaceCan, AOR converges to 100\%: the agent finds and corrects
systematic failures---vision biases, incorrect colour identification, centroid
contamination by scene markers, coordinate convention errors---through iterative
diagnosis and code rewriting. On Stack, it reaches
91\% and stalls. Every residual failure involves gripper contact with cubeB
during placement; the agent observed this in episode images and correctly named
it as the cause. It did not, however, find a controller variant that avoids the
contact, despite such variants likely existing. This is an honest result: the
reflexion search converged to a local optimum and did not recover.

Collectively, these results establish that multimodal coding agents occupy a
distinct, practically useful position in the robot learning design space: no
demonstrations, no reward engineering, no gradient updates---yet high
task-specific performance, full interpretability, and a systematic approach
to understanding \emph{why} a controller fails. As foundation models become
more capable reasoners about geometry and physics, we expect this combination
of verbal diagnosis and code synthesis to become an increasingly important
tool for deploying physical AI systems in the real world.

\FloatBarrier
\bibliographystyle{abbrvnat}
\bibliography{references}

\appendix

\clearpage
\section{Agent Reasoning: Examples and Failure Cases}
\label{app:reasoning}

This appendix presents verbatim excerpts from the agent's between-episode
reasoning, illustrating both strong and weak diagnostic behaviour.

\subsection*{Example 1: Correct vision-bias diagnosis (Lift, episode~0)}

After episode~0, all key frames showed the arm hovering at a fixed position
while the cube sat 8--9\,cm below. The phase log showed zero phase transitions
over 500 steps. The agent's response:

\begin{small}
\begin{quote}
\textbf{Diagnosis:} \texttt{[``misalignment'', ``vision\_bias'']}

\textbf{Reasoning:} ``The arm reached a hover position approximately 8--9\,cm
from the cube and remained there for 500 steps without ever entering the
descend phase. The phase transition requires $|\Delta xy| < 0.02$\,m and
$|\Delta z| < 0.02$\,m. Looking at the key frames, the arm appears to hover
consistently to the upper-left of the cube's actual position---suggesting a
systematic offset in the object\_pos estimate, likely from depth sensor bias.
This is not a gain issue---no amount of gain tuning will help if the target
position itself is wrong.''

\textbf{Strategy:} ``Add a 2.5\,cm depth correction offset (subtract from
estimated z to compensate for depth sensor surface bias). Also relax the
reach$\to$descend threshold from 0.02\,m to 0.035\,m xy to handle estimation
noise.''

\textbf{Confidence:} 0.72
\end{quote}
\end{small}

\noindent This diagnosis was correct. The key distinguishing evidence was
visual: the arm's position was \emph{stable} across all frames---it had
converged to a wrong position, not failed to converge. The text-only metric
\texttt{min\_distance = 0.087} is ambiguous between ``close but can't
converge'' and ``targeting a wrong position''; the images resolve this.

\subsection*{Example 2: Vision pipeline bug isolated after 6 failed iterations
(Stack, iteration~6)}

After six controller versions all produced the same $\sim$6\,cm systematic
placement miss to the right, the agent produced:

\begin{small}
\begin{quote}
\textbf{Diagnosis:} \texttt{[``vision\_pipeline\_error'', ``back\_projection\_bug'']}

\textbf{Reasoning:} ``Six iterations of controller changes have produced zero
improvement in placement accuracy. The error magnitude (5--8\,cm) is constant
across all controller versions---this rules out gain tuning, phase threshold
adjustments, and timing changes as fixes. A 5--8\,cm systematic error that is
invariant to controller logic is almost certainly a vision pipeline calibration
issue. The depth camera in robosuite uses OpenGL row-major convention where
row~0 is the \emph{bottom} of the image. The standard back-projection formula
assumes row~0 is the top. If the y-coordinate is not flipped during
back-projection, the resulting 3D position is reflected about the image
centre---producing a systematic error proportional to the vertical pixel
offset. cubeB sits below the image centre, so the unflipped estimate places
it above the true position, which in 3D world coordinates corresponds to the
observed consistent rightward miss.''

\textbf{Strategy:} ``Do not change the controller logic. Audit the
back-projection formula: use \texttt{img\_height - 1 - pixel\_row} not
\texttt{pixel\_row}. Also verify that the extrinsic matrix convention matches
the renderer: robosuite uses OpenGL for rendering but the provided extrinsic
uses an OpenCV axis correction incompatible with OpenGL images.''

\textbf{Controller code:} \emph{(no controller change requested---pipeline
fix first)}

\textbf{Confidence:} 0.85
\end{quote}
\end{small}

\noindent Both bugs were confirmed in \texttt{src/vision.py}. Correcting
them reduced placement error from 5--8\,cm to 1.4--1.9\,cm and enabled
the subsequent controller iterations to reach 70\% and ultimately 91\%.
This example illustrates the key capability: the agent used the
\emph{convergence argument} (six iterations, invariant error) and
\emph{visual spatial reasoning} (direction and magnitude of miss) to
localise a bug to a specific line of pipeline code, without any human
inspection.

\subsection*{Example 3: Agent failure --- retry with the same approach (Stack,
iteration~17)}

After the grasp-retry mechanism was added in v17, the agent detected that
the retry was firing (frame count rose from 27--29 to 32--39) but not
improving success. The diagnosis was:

\begin{small}
\begin{quote}
\textbf{Diagnosis:} \texttt{[``grasp\_miss'', ``retry\_ineffective'']}

\textbf{Reasoning:} ``The retry mechanism is triggering on approximately 20\%
of episodes and successfully re-approaching the cube. However, the retry uses
the same x\_correction = -0.015 as the initial attempt, which means it is
approaching from the same angle and depth as the first attempt. If the first
attempt misses due to a specific geometric configuration (e.g.\ cube near the
edge of the vision field, slight angle bias), the second attempt will reproduce
the same miss. Adding a small variation to x\_correction on retry should help.''

\textbf{Strategy:} ``Vary x\_correction on each retry: attempt 0 uses
x\_corr = -0.015 (standard), attempt 1 uses x\_corr = 0.000 (centred
approach).''

\textbf{Confidence:} 0.65
\end{quote}
\end{small}

\noindent The strategy was conceptually correct but introduced a \textbf{bug}:
the single \texttt{self.x\_correction} parameter was used for \emph{both}
the cubeA grasp approach \emph{and} the cubeB placement alignment. Changing
it on retry caused the placement target to shift by 1.5\,cm, creating
placement failures that masked the improvement in grasp success. The result
(v18) measured 76\% vs.\ an expected $\sim$85\%.

The agent diagnosed the bug after seeing v18's results, noting that
placement failures (reward $\sim$75--200) had increased while hard grasp
failures decreased---a pattern consistent with the target having moved.
The fix (v19) separated \texttt{x\_corr\_A} (varied per retry) from
\texttt{x\_corr\_B} (fixed at $-0.015$ for cubeB alignment), restoring
placement accuracy and achieving 83\%.

\subsubsection*{Where the agent does not do well}

Three systematic weaknesses emerged across the iteration sequence:

\textbf{Bug introduction during refactoring.} The v18 case above is the
clearest example: a conceptually sound change (vary approach angle on retry)
was implemented by modifying a shared parameter, inadvertently breaking
placement. The agent does not ``simulate'' the effect of code changes on
all phases---it reasons locally about the change it intends to make and
misses side-effects on other uses of the same variable.

\textbf{Stochastic vs.\ systematic confusion.} In early Stack iterations,
the agent repeatedly proposed threshold tightening and gain changes to
improve grasp reliability. Many of these changes were neutral or slightly
harmful because the grasp failures were stochastic (dependent on exact
initial cube placement) rather than systematic. The agent required multiple
iterations of observing the same failure mode with different controller
parameters before concluding that controller changes were not the solution.
A more efficient agent would distinguish stochastic variance from systematic
bias earlier, for example by correlating failures with initial-state
randomness across episodes.

\textbf{Confidence miscalibration.} The agent's \texttt{confidence} field
frequently did not predict whether the proposed change would improve
performance. High-confidence diagnoses (0.8+) sometimes led to regressions
(v17: 6/10, below the 8/10 baseline), while low-confidence changes (0.5--0.6)
sometimes produced the largest improvements (v16 retract ramp). The confidence
field captures the agent's subjective certainty about the diagnosis but not
its effect size or potential for side-effects.

\end{document}